# Non-Existence of Linear Universal Drift Functions

Benjamin Doerr   Daniel Johannsen
Carola Winzen
Max-Planck-Institut für Informatik
Campus E1 4
66123 Saarbrücken, Germany

**Abstract**

Drift analysis has become a powerful tool to prove bounds on the runtime of randomized search heuristics. It allows, for example, fairly simple proofs for the classical problem how the (1+1) Evolutionary Algorithm (EA) optimizes an arbitrary pseudo-Boolean linear function. The key idea of drift analysis is to measure the progress via another pseudo-Boolean function (called drift function) and use deeper results from probability theory to derive from this a good bound for the runtime of the EA. Surprisingly, all these results manage to use the same drift function for all linear objective functions.

In this work, we show that such universal drift functions only exist if the mutation probability is close to the standard value of $1/n$.

## 1 Introduction

An innocent looking problem is the question how long the (1+1) pseudo-Boolean Algorithm ((1+1) EA) needs to find the optimum of a given linear function. However, this is in fact one of the problems that was most influential for the theory of evolutionary algorithms.

While particular linear functions like OneMax were easily analyzed, it took a major effort by Droste, Jansen and Wegener [DJW02] to solve the problem in full generality and to show that the (1+1) EA optimizes any linear function in $O(n \log n)$ steps. Their proof of the result, however, is highly technical.

A major breakthrough spurred by this problem is the work by He and Yao [HY01, HY03], who introduced *drift analysis* to the field of evolutionary computation. This allowed a significantly simpler proof for the linear functions problem. Even more important, drift analysis quickly became one of the most powerful tools for both proving upper and lower bounds on the runtime of evolutionary algorithms. See, e.g., [HY03, GL06, HJKN08, OW10, He10].

In a nut-shell, the drift analysis conducted by He and Yao is a potential function argument. For a suitable potential function (usually called *drift function*), they show that in each iteration of a run of the (1+1) EA optimizing the original linear function, also a certain improvement with respect to the drift function is obtained. By this, stopping-time arguments which were difficult to obtain for the original function can be replaced by such arguments for the drift function.

However, the proof given by He and Yao [HY01, HY03] is still not easy. The difficulties include both finding a suitable drift function and proving that this function has a positive drift in every search point.

Another great progress was made by Jägerskü̈pper [Jäg08], who used a clever averaging argument avoiding the need to show that from each search point on there is a positive drift. In consequence, he was able to use as drift function the natural ONEMAX function, which simply counts the number of 1-bits in the bit string. This also allowed to determine reasonable values for the usually not explicitly given constants. More precisely, Jägerskü̈pper showed that the expected optimization time for any linear function defined on $n$-bit strings is bounded by $(1 + o(1))2.02en \ln(n)$.

In [DJW10b], a multiplicative drift theorem was proposed. It allows a simpler and more natural proof of the $O(n \log n)$ bound. By combining Jägerskü̈pper's arguments from [Jäg08] with the multiplicative drift theorem, the authors improved his upper bound to $(1 + o(1))1.39en \ln(n)$.

Interestingly, in each of these results the same drift function could be used for all linear objective functions. We call such a drift function *universal*.

## Our Results

In this work, we show that the existence of universal drift functions depends highly on the mutation probability. If this is larger than the standard value of $1/n$ by more than a certain constant factor, universal drift functions do not exist. In consequence, it is not clear how to extend previous results to large mutation probabilities.

We show that the $(1 + 1)$ $EA_c$ does not allow linear universal drift functions even for relatively small values of $c$. More precisely, we show that the classical, additive drift method by He and Yao does not allow universal drift functions for values of $c$ larger than 4. The multiplicative drift method does not allow linear universal drift functions for values of $c$ greater then 2.2. Lastly, we show that even if we combine the Jägerskü̈pper approach with the multiplicative drift method, linear universal drift functions do not exist if the mutation probability exceeds $7/n$.

## 2 Optimizing Linear Functions with the (1+1) EA

Throughout this paper, we are interested in the performance of different variants of the (1+1) EA on the class of linear functions. To be more precise, we are interested in the proof techniques which allow us to show a certain behavior of the algorithms on this class of functions.

The aim of this section is to give a very brief introduction to the class of linear functions and to the algorithms under consideration.

Before we move on, here is some notation which we use throughout the work. By $\mathbb{N}$ we denote the set of positive integers and, accordingly, we set $\mathbb{N}_0 := \mathbb{N} \cup \{0\}$. If no further comments are made, $n$ will always denote the length of the input, i.e., in our case, the length of the bit strings in the considered search space. For convenience, we write $[n]$ for $\mathbb{N}_{\leq n}$. A bit string $x \in \{0,1\}^n$ is denoted by $x_n \ldots x_1$. This notation is inspired by the function BINVAL (which will be defined below). It allows us to use the standard notation for binary representation of natural numbers.

For every $i \in \mathbb{N}_{\leq n}$ let $e_i \in \{0,1\}^n$ be the $i$-th unit vector, i.e., $(e_i)_j = 1$ if and only if $j = i$. By $\oplus$ we denote the bitwise XOR operation on bit strings, i.e., for all $x, y \in \{0,1\}^n$ we have $(x \oplus y)_i = 1$ if and only if $x_i \neq y_i$. For a stochastic event $A$, we denote by $\chi(A)$ the characteristic function of $A$, i.e., $\chi(A) = 1$ if $A$ occurs, and $\chi(A) = 0$ otherwise.

### 2.1 Linear Functions

**Definition 1** (Linear Functions). *Let $n \in \mathbb{N}$. A function $f: \{0,1\}^n \to \mathbb{R}$ is called* linear *if there exist weights $w_1, \ldots, w_n \in \mathbb{R}$ such that for all $x \in \{0,1\}^n$ it holds that $f(x) = \sum_{j=1}^n w_j x_j$. The class of linear functions* LINEAR *is defined as the set of all such functions, i.e.,*

$$\text{LINEAR} := \{f : \{0,1\}^n \to \mathbb{R}, x \mapsto \sum_{j=1}^n w_j x_j \mid w_1, \ldots, w_n \in \mathbb{R}\}. \quad (1)$$

We say that $f$ has *monotone weights* if $w_1 \leq \cdots \leq w_n$. It is easy to see (and has been argued in [DJW02]) that when analyzing how the (1+1) EA optimizes a linear function we can assume monotone weights without loss of generality. Furthermore, we always assume $w_1 > 0$, again not restricting the generality of the results.

For the purpose of better legibility, we ignore the dilemma of using $f$ as name for the function itself and for its weights and write $f(x) = \sum_{j=1}^n f_j x_j$ and, similarly, $g(x) = \sum_{j=1}^n g_j x_j$ for linear functions $f$ and $g$.

Also note that in this paper we are interested in the number of iterations it takes to *minimize* a given linear function. Note, however, that the optimization time bounds obtained in this work are the same for the minimization and the maximization problem.

```
1  Choose x^(0) ∈ {0,1}^n uniformly at random;
2  for t = 1 to ∞ do
3  |   Sample y^(t) ∈ {0,1}^n by flipping each bit in x^(t-1) with
   |   probability c/n;
4  |   if f(y^(t)) ≤ f(x^(t-1)) then
5  |   |   x^(t) := y^(t)
6  |   else
7  |   |   x^(t) := x^(t-1)
```

**Algorithmus 1:** $(1+1)$ $EA_c$: The (1+1) Evolutionary Algorithm with mutation rate $c/n$ for minimizing $f\colon \{0,1\}^n \to \mathbb{R}$.

In the following discussions, two linear functions will play an important role. The first one, the so-called ONEMAX function, simply counts the number of ones in the bit string, i.e., $\text{ONEMAX}(x) = \sum_{j=1}^{n} x_j$. We shall often abbreviate $\text{ONEMAX}(x)$ by $|x|_1$, in particular if the space is limited. The second function of particular interest is BINVAL. It is defined by $\text{BINVAL}(x) = \sum_{j=1}^{n} 2^j x_j$. We will discuss some properties of these two functions in the next subsection.

## 2.2 The (1+1) Evolutionary Algorithm

Our main interest in this work is to show that certain analysis techniques for the (1+1) Evolutionary Algorithm fail if we change the mutation probability only by a constant factor. We denote by (1+1) $EA_c$ the (1+1) EA where the standard mutation rate of $p = 1/n$ is changed to $p = c/n$, where $c$ is some positive constant (cf. Algorithm 1).

Let us comment on some features of the (1+1) $EA_c$ as described in Algorithm 1. It starts with a randomly chosen initial bit string $x^{(0)}$. Thus, on average, we expect $n/2$ bits to be 0 and the other half to equal 1. In each iteration $t \geq 1$ the (1+1) EA performs two steps.

The *mutation step* can be described as follows: The algorithm creates a random vector $Y^{(t)} \in \{0,1\}^n$ such that $\Pr[Y_i^{(t)} = 1] = c/n$ mutually independent for all $i$. Then, $\tilde{x}^{(t-1)} = x^{(t-1)} \oplus Y^{(t)}$ is the new candidate for the next search point.

In the *selection step*, the algorithm ensures that $\tilde{x}^{(t-1)}$ is accepted as a new search point only if it is at least as good as the current solution, i.e.,

$$x^{(t)} = \begin{cases} \tilde{x}^{(t-1)} & \text{if } f(\tilde{x}^{(t-1)}) \leq f(x^{(t-1)}), \\ x^{(t-1)} & \text{otherwise.} \end{cases}$$

The expected number of iterations $T$ until the (1+1) $EA_c$ selects for the first time a bit string $x$ such that $f(x)$ is minimal is called the *optimization* *time* of the (1+1) EA$_c$. Note that one iteration consists of exactly one mutation and one selection step.

Let us consider the behavior of the (1+1) EA, on our two example functions, ONEMAX and BINVAL. As ONEMAX simply counts the number of 1s in the bit string, the the following holds. If during one iteration of the (1+1) EA the string $y$ is created from $x$, $y$ is accepted as a new search point for the next iteration if and only if $|y|_1 \leq |x|_1$.

The situation is different for the second example function, BINVAL. When optimizing this function, the inequality $2^i > \sum_{j=1}^{i-1} 2^j$ implies that the algorithm accepts a new bit string if and only if the highest-indexed bit that is touched by the mutation is flipped from one to zero.

In spite of this different behavior, Droste, Jansen and Wegener could prove in their seminal paper [DJW02] that the expected optimization time of the standard (1+1) EA (with mutation probability $p = 1/n$) is of the order $(n \log n)$ for all linear functions.

**Theorem 2** ([DJW02]). *The expected optimization time of the standard (1+1) EA on any linear function with positive weights is $\Theta(n \log n)$.*

A more precise upper bound of $(1 + o(1))2.02en \ln n$ was provided by Jägerskükpper [Jäg08]. In [DJW10a], the authors of this paper improved the bound to $(1 + o(1))1.39en \ln n$, together with a lower bound of $(1 - o(1))en \ln n$. A discussion of the proof methods is given in the next section.

## 3  Drift Analysis

Drift analysis has been introduced to the field of evolutionary computation by He and Yao [HY01, HY03]. The method builds on a result of Hajek [Haj82]. The main idea of He and Yao is the following. When analyzing the optimization behavior of a randomized search heuristic, instead of tracking how the objective function becomes better, one uses an auxiliary function, the so-called *potential* or *drift function* and tracks its behavior. The drift function is typically designed in such a way that it is minimal if and only if the objective function itself is minimized. We give an example after the formal description of the method.

### 3.1  Additive Drift

The following additive drift theorem was introduced to the field of evolutionary computation by He and Yao.

**Theorem 3** (Additive Drift Theorem [HY04]). *Let $\{Z^{(t)}\}_{t \in \mathbb{N}_0}$ be random variables describing a Markov process over a finite state space $\mathcal{S} \subseteq \mathbb{R}$. Let $T$ be the random variable that denotes the earliest point in time $t \in \mathbb{N}_0$ such that $Z^{(t)} \leq 0$.*

*If there exist $\delta > 0$ and $c > 0$ such that*

(i) $E[Z(t) - Z(t+1)|t < T] \geq \delta$ and

(ii) $Z^{(0)} \leq c$.

Then $E[T] \leq \frac{c}{\delta}$.

The idea of applying this theorem to the analysis of the (1+1) EA is as follows. Given a function $f$ and a mutation probability $p = c/n$, let us denote by $\{x^{(t)}\}_{t \in \mathbb{N}_0}$ the (random) series of the search points (after selection) of one run of the (1+1) $EA_c$. We now try to find another function $g$ such that

(a) $\{x \mid f(x) \text{ minimal}\} = \{x \mid g(x) \text{ minimal}\}$ and

(b) $\{Z^{(t)}\}_{t \in \mathbb{N}_0} := \{g(x^{(t)})\}_{t \in \mathbb{N}_0}$ fulfills the requirements of Theorem 3.

The drift theorem then provides an upper bound for the expected time needed by the (1+1) $EA_c$ to minimize $g$. Condition (a) ensures that the same upper bound holds for $f$ as well.

Condition (b) is typically a little tricky to prove. It requires that, given some $x \in \{0,1\}^n$, we can expect, on average, that $g$ becomes smaller whenever $f$ does. That is, $g$ is *drifting* towards the same direction as the objective function $f$ itself. That is why we call $g$ a drift function for $f$.

Let us give a short example. When using the (1+1) $EA_1$[1] to minimize a pseudo-Boolean linear function $f \colon \{0,1\}^n \to \mathbb{R}, x \mapsto \sum_{j=1}^n f_j x_j$ with arbitrary positive weights $0 < f_1 \leq \cdots \leq f_n$, the drift function can be chosen as $g \colon \{0,1\}^n \to \mathbb{R}, x \mapsto \ln\left(1 + \sum_{j=1}^{\lfloor n/2 \rfloor} x_j + \sum_{j=\lfloor n/2 \rfloor+1}^n 2\, x_j\right)$.

Though still needing some calculations, one can show the following. If $y$ is the result of one iteration (mutation and selection) of the (1+1) EA starting in some non-optimal $x \in \{0,1\}^n$, then

$$E[g(x) - g(y)] \geq \delta/n \tag{2}$$

for some $\delta > 0$. The application of Theorem 3 yields that after an expected number of $g(x)/(\delta/n) = O(n \log n)$ iterations, our initial $g$–value of $g(x)$ is reduced to 0. But $g(y) = 0$ implies $f(y) = 0$, that is, the (1+1) EA has found the desired optimum of $f$ after $O(n \log n)$ iterations.

### 3.2 Multiplicative Drift

Using drift analysis usually bears two difficulties. The first is guessing a suitable drift function $g$. The second, related to the first, is proving that during a run of the (1+1) EA, $f$ and $g$ behave sufficiently similar, that is, we can prove some statement like inequality (2). Note that this inequality contains information about $f$ as well, namely implicitly in the fact that $y$ has an at least as good $f$–value as $x$.

---

[1] In fact, He and Yao analyzed a variant of the (1+1) EA presented here. In this variant, a candidate search point is only accepted if it is strictly better than the current optimum. However, the results for upper bounds in this work can easily be transfered to the setting of [HY01, HY03].

What makes showing that $g$ (as chosen in Subsection 3.1) is a suitable drift function particularly costly, is the logarithm around the simple linear function $\tilde{g}(x) = \sum_{j=1}^{\lfloor n/2 \rfloor} x_j + \sum_{j=\lfloor n/2 \rfloor+1}^{n} 2\, x_j$.

This motivated the authors to formulate a different, multiplicative version of He and Yao's drift method in [DJW10b]. Although the method can easily be derived from the additive version, it has been shown to be a very natural and elegant way for proving upper bounds on the optimization time of randomized search heuristics on different classes of problems.

**Theorem 4** (Multiplicative Drift Theorem [DJW10b])**.** *Let $S \subseteq \mathbb{R}^+$ be a finite set with minimum $s_{\min} := \min\{s \in S\}$. Let $\{X^{(t)}\}_{t\in\mathbb{N}}$ be a sequence of random variables over $S \cup \{0\}$. Let $T$ be the random variable that denotes the first point in time $t \in \mathbb{N}$ for which $X^{(t)} = 0$.*

*Suppose that there exists a constant $\delta > 0$ such that*

$$\mathrm{E}\left[X^{(t)} - X^{(t+1)} \mid X^{(t)} = s, T > t\right] \geq \delta s \tag{3}$$

*holds for all $s \in S$ such that $\Pr[X^{(t)} = s, T > t] > 0$. Then, for all $s_0 \in S$,*

$$\mathrm{E}\left[T \mid X^{(0)} = s_0\right] \leq \frac{1 + \ln(s_0/s_{\min})}{\delta}.$$

Note that, whenever $g$ is a drift function for some function $f$ in the sense that

(i) $\{x \mid f(x) \text{ minimal}\} = \{x \mid g(x) \text{ minimal}\}$ and

(ii) $\{X^{(t)}\}_{t\in\mathbb{N}_0} := \{g(x^{(t)})\}_{t\in\mathbb{N}_0}$ fulfills the requirements of Theorem 4,

the function $\ln(1+g)$ is a drift function in the classical sense of Section 3.1. The opposite direction is not true. I.e., if $g$ is a linear function such that $\ln(1 + g)$ is a drift function for $f$ in the sense of Theorem 3, one cannot conclude that $g$ itself is a drift function in the multiplicative setting of Theorem 4. However, the analyses carried out in Section 4 can be transferred to the additive setting, as shown in Subsection 4.2.

Let us note that it has been shown in [DJW10b] that the multiplicative drift theorem allows a fairly simple proof for Theorem 2. There, the drift function is chosen to be $g \colon \{0,1\}^n \to \mathbb{R}, x \mapsto \sum_{j=1}^{n} g_j x_j$ with $g_j = 1$ for $j \leq n/2$ and $g_j = 5/4$ otherwise.

## 3.3 Distribution-based Multiplicative Drift

One may ask why, in the additive setting, not to take $g(x) := \ln(1 + \textsc{OneMax}(x))$ as potential function. However, an easy observation reveals that there is an objective function $f$ and a search point $x$ such that $g$ yields to small a drift with respect to $f$. To see this, let $x := e_n = (1, 0, \ldots, 0)$ and let $f := \textsc{BinVal}$ be the function to be optimized. Then the point-wise drift (2) of $g$ is only of order $\Theta(1/n^2)$. This example shows that finding a drift function yielding point-wise drift for all $x$ and all $f$ is not so easy.

The first to overcome the difficulties of point-wise drift was Jägersküpper [Jäg08]. While he still avoids completely analyzing the actual distribution of $x^{(t)}$, he does show the following property of this distribution which in turn allows him to use a distibution-based drift approach. In this way, he omits the need for point-wise drift. Jägersküpper's simple observation is that at any time step $t$, the more valuable bits are more likely to be in the desired setting.

**Theorem 5** ([Jäg08]). *Let $n \in \mathbb{N}$ and let $x^{(t)}$ denote the random individual (distributed over $\{0,1\}^n$) after $t \in \mathbb{N}_0$ iterations of the (1+1) $EA_1$ minimizing a linear function $f \colon \{0,1\}^n \to \mathbb{R}$. Then,*

$$\Pr[x_1^{(t)} = 0] \leq \cdots \leq \Pr[x_n^{(t)} = 0].$$

*This statement remains true if we condition on $|x^{(t)}|_1 = k$ for any $k \in [n]$.*

Using this theorem, Jägersküpper was able to show a lower bound of $\Omega(1/n)$ for the multiplicative drift of $\textsc{OneMax}$ as potential function for any linear function.

**Proposition 6.** *Let $n \in \mathbb{N}$, let $f \colon \{0,1\}^n \to \mathbb{N}_0$ be linear and let $g := \textsc{OneMax}$. Let $x^{(t)}$ be the individual in the t-th iteration of the (1+1) $EA_1$ minimizing $f$. Then,*

$$\mathrm{E}[g(x^{(t)}) - g(x^{(t+1)}) \mid g(x^{(t)}) = k] \geq \frac{(e-2)k}{en}.$$

*holds for all $k \in \mathbb{N}$ and $t \in \mathbb{N}_0$ with $\Pr[g(x^{(t)}) = k] > 0$.*

Proposition 6 shows that it is even possible to take $\textsc{OneMax}$ as a drift function if we consider the $(1+1)$ $EA_1$. Using this approach, Jägersküpper was not only able to give a more natural proof for the $O(n \log n)$ bound of the (1+1) $EA_1$ on the class of linear functions, but he could also give a meaningful upper bound on the leading constant. More precisely, he shows that the expected optimization time of the (1+1) EA minimizing a linear function on $n$ bits is at most $(1 + o(1))2.02en \ln(n)$.

# 4 Non-Existence of Linear Universal Drift Functions

In the previous section, we have seen for the different drift methods that, if we are considering the standard (1+1) $EA_1$ with mutation probability $1/n$, we are able to define a linear function $g$ such that $g$ (or $\ln(1+g)$, respectively) serves as a good drift function for all linear functions, independently of the particular weights. In the following, we call such functions $g$ linear universal drift functions. We give a more precise definition below.

In this section we prove the main results of this paper, namely that universal linear drift functions with monotone weights do not exist if the mutation probability exceeds $c/n$ for some small, setting-dependent constant $c$.

Before we formulate the theorems, let us introduce the operator $\Delta_c$ which measures the progress made by the (1+1) $EA_c$ on $f$ with respect to some other function $g$.

**Definition 7** ($\Delta_c(g,f,x)$)**.** *Let $Y \in \{0,1\}^n$ be randomly chosen such that $\Pr[Y_i = 1] = c/n$ mutually independent for all $i \in [n]$. For $f$ and $g\colon \{0,1\}^n \to \mathbb{R}$ and for $x \in \{0,1\}^n$ we define the random variable $\Delta_c(g,f,x)$ by*

$$\Delta_c(g,f,x) := (g(x) - g(x \oplus Y)) \cdot \chi(f(x \oplus Y) \leq f(x)).$$

If we are considering the multiplicative setting from Subsections 3.2 and 3.3, we say that $g$ is a *linear universal drift function*, if $g$ itself is linear and if $\Delta_c(g,f,x) \geq 0$ for all linear functions $f$ with monotone weights and all possible search points $x \in \{0,1\}^n$. When we consider the additive setting from 3.1, the second condition translates to $\Delta_c(\ln(1+g), f, x) \geq 0$.

A definition for the distribution-based setting of Jägerskupper will be given in Subsection 4.3.

## 4.1 Multiplicative Setting

We first show the non-existence result for the multiplicative setting. Intuitively speaking, it tells us that linear universal drift functions do not exist if the mutation probability is larger than $2.2/n$. We then present in the next subsection how this result can be transferred to the setting of the additive drift theorem.

**Theorem 8** (Nonexistence Theorem for Multiplicative Drift)**.** *Let $n \in \mathbb{N}$ be sufficiently large and let $c > 2.2$. If we consider the (1+1) $EA_c$, the following statement holds. For every linear function $g\colon \{0,1\}^n \to \mathbb{R}, x \mapsto \sum_{j=1}^n g_j x_j$ with $1 = g_1 \leq \cdots \leq g_n$, there exist a linear function $f$ with monotone weights and a bit string $x \in \{0,1\}^n$ such that $\mathrm{E}[\Delta_c(g,f,x)] < 0$.*

We prove the theorem by contraposition. To this end, let $n \in \mathbb{N}$ be sufficiently large and let us assume that there exists a universal linear drift function $g$ with $1 = g_1 \leq \cdots \leq g_n$. That is, for every linear function $f$ with monotone weights and every $x \in \{0,1\}^n$ we have $\mathrm{E}[\Delta_c(g, f, x)] \geq 0$. It suffices to show that $c$ cannot be larger than 2.2.

The proof is structured as follows: In Proposition 9 and Corollary 10 we derive lower bounds for $\sum_{j=1}^n g_j$. An upper bound is given in Proposition 11. The combination of the three results will conclude the proof.

**Proposition 9.** *If we consider the (1+1) $EA_c$ for some constant $c$ and if $g$ is a universal linear drift function with monotone weights, then for all $1 \leq i \leq n$ we have $g_i \geq \frac{c}{n} \sum_{j=1}^{i-1} g_j$.*

*Proof.* Let $f = \text{BinVal}$, $i \in [n]$ and $x = e_i$. Let $A$ be the event that the $i$th bit is the smallest indexed flipping bit. Formally, let $Y \in \{0,1\}^n$ be vector indicating which bits are being flipped, i.e., $Y_i = 1$ if and only if the $i$-th bit $x_i$ of $x$ is flipped. Then event $A$ happens if and only is $Y_i = 1$ and $Y_j = 0$ for all $j > i$. Clearly, $A$ expresses the event that $x \oplus Y$ is accepted as the new search point and $x \oplus Y \neq x$. That is, $\chi(f(x \oplus Y) \leq f(x)) = \chi(A)$. Thus,

$$\mathrm{E}[\Delta_c(g, f, x)] = \mathrm{E}[g(x) - g(x \oplus Y) \mid A] \cdot \Pr[A].$$

It is easy to verify that

$$\Pr[A] = \tfrac{c}{n}(1 - \tfrac{c}{n})^{n-i},$$

which is strictly positive. From $0 \leq \mathrm{E}[\Delta_c(g, f, x)]$ we conclude

$$0 \leq \mathrm{E}[g(x) - g(x \oplus Y) \mid A] = g_i - \sum_{j=1}^{i-1} \tfrac{c}{n} g_j$$

and the statement follows. □

**Corollary 10.** *Let us consider the (1+1) $EA_c$ for some constant $c > 1$ and let $g$ be a universal linear drift function with monotone weights. For $k := \lceil \frac{n}{c} \rceil$ and $\ell \in \{1, ..., n-k\}$ it holds that*

$$g_{k+\ell} \geq (1 + \tfrac{c}{n})^{\ell-1}.$$

*Proof.* We show the claim via induction with respect to $\ell$. By definition, $g_{k+1} \geq g_1 = 1$. Now, for $\ell \geq 1$, Proposition 9 and the induction hypothesis yield

$$g_{k+\ell+1} \geq \tfrac{c}{n}\left(\sum_{j=1}^k 1 + \sum_{j=k+1}^{k+\ell} g_j\right) \geq 1 + \tfrac{c}{n}\sum_{j=1}^\ell (1+\tfrac{c}{n})^{j-1} = 1 + \tfrac{c}{n}\,\frac{(1+\tfrac{c}{n})^\ell - 1}{\tfrac{c}{n}}$$

and again the statement follows. □

We now prove an upper bound for the sum of the weights of $g$.

**Proposition 11.** *Let $c$ a constant and let $p := c/n$. If we consider the (1+1) $EA_c$ and if $g$ is a linear universal drift function with monotone weights, it holds that $\sum_{j=1}^n g_j \leq \frac{1+np-p}{p}$.*

*Proof.* Let $f = \textsc{OneMax}$ and $x = e_1$. Then, $f(x) = 1$ and the event $f(x \oplus Y) \leq f(x)$ occurs if and only if

$$f(x \oplus Y) = \sum_{j=1}^n (x \oplus Y)_j \leq 1.$$

Therefore, let us denote by $A$ be the event that $Y = e_1$, by $B_j$ the event that $Y = e_1 \oplus e_j$ for $j > 1$. Finally, let us denote by $C$ the event that $Y = 0$. Then,

$$\chi\big(f(x \oplus Y) \leq f(x)\big) = \chi(A) + \sum_{j=2}^n \chi(B_j) + \chi(C).$$

Thus,

$$\begin{aligned}
\mathrm{E}[\Delta_c(g,f,x)] &= \mathrm{E}[g(x) - g(x \oplus Y) \mid A] \cdot \Pr[A] \\
&+ \sum_{j=2}^n \mathrm{E}[g(x) - g(x \oplus Y) \mid B_j] \cdot \Pr[B_j] \\
&+ \mathrm{E}[g(x) - g(x \oplus Y) \mid C] \cdot \Pr[C],
\end{aligned}$$

the latter summand equaling 0. Now,

$$\begin{aligned}
\mathrm{E}[g(x) - g(x \oplus Y) \mid A] &= g_1, \\
\mathrm{E}[g(x) - g(x \oplus Y) \mid B_j] &= g_1 - g_j, \\
\Pr[A] &= (1-p)^{n-1} p, \text{ and} \\
\Pr[B_j] &= (1-p)^{n-2} p^2.
\end{aligned}$$

Since $g$ is a drift function for $f$ we have $\mathrm{E}[\Delta_c(g,f,x)] \geq 0$. Hence,

$$0 \leq \mathrm{E}[\Delta_c(g,f,x)] = (1-p)^{n-2} p \left( (1-p)g_1 + p \sum_{j=2}^n (g_1 - g_j) \right),$$

yielding

$$0 \leq (1-p)g_1 + p \sum_{j=2}^n (g_1 - g_j).$$

By resorting we have

$$p \sum_{j=1}^n g_j \leq (1 + (n-1)p)g_1 = 1 - p + np,$$

which concludes the proof. $\square$

The upper and lower bounds for the weights of $g$ now allow us to prove Theorem 8:

*Proof of Theorem 8.* We need to prove $c \leq 2.2$. Let us again abbreviate $p := c/n$. For the purpose of better readability let $k := \lceil \frac{1}{p} \rceil$. Propositions 9 and 11 yield

$$\frac{1-p+np}{p} \geq \sum_{i=1}^{n} g_i \geq k \cdot 1 + \sum_{i=0}^{n-k-1} (1+p)^i = k + \frac{(1+p)^{n-k}-1}{p}$$
$$\geq \frac{(1+p)^{n-k}}{p}.$$

Thus,
$$1 \geq (1+p)^{n-k} + p(1-n). \tag{4}$$

By re-substituting $p$ with $c/n$, the term on the right-hand side can be bounded from below by

$$\left(1+\tfrac{c}{n}\right)^{n-\tfrac{n}{c}-1} + \tfrac{c}{n} - c = \left(1+\tfrac{c}{n}\right)^{n(1-\tfrac{1}{c}-\tfrac{1}{n})} + \tfrac{c}{n} - c$$
$$= e^{c-1}(1-o(1)) - c$$

For sufficiently large $n$ and $c > 2.2$, this term exceeds 1, which contradicts inequality (4). Hence, $c \leq 2.2$. □

## 4.2 Additive Setting

We now transfer the results obtained in the previous subsection to the additive setting. That is, we are interested in the question "Can we find a linear function $g$ such that $\ln(1+g)$ serves as a drift function for all linear functions (with monotone weights)?". We can apply the methodology of the previous propositions to show that such a linear universal function $g$ does not exist if the mutation probability $c/n$ is large.

We do not try to find the best possible constant, but prefer to use the simple approach obtained via the multiplicative drift in the previous subsection. We then use a numerical example to show that linear universal drift functions do not exist if $n = 100$ and $c \geq 4$.

**Theorem 12.** *Let $n = 100$ and, $c \geq 4$ and let us consider the (1+1) $EA_c$. For every linear function $g\colon \{0,1\} \to \mathbb{R}, x \mapsto \sum_{j=1}^{n} g_j x_j$ with weights $1 = g_1 \leq \cdots \leq g_n$ there exists an $x \in \{0,1\}^n$ and a linear function $f$ with monotone weights such that we have $E[\Delta_c(\ln(1+g), f, x)] < 0$.*

We are going to use the tools that we have just developed for the multiplicative setting. Thus, we again apply contraposition. Therefore, let us fix some function $g\colon \{0,1\} \to \mathbb{R}, x \mapsto \sum_{j=1}^{n} g_j x_j$ with $1 = g_1 \leq \cdots \leq g_n$ such that $E[\Delta_c(\ln(1+g), f, x)] \geq 0$ for all $f$ and all $x$ as in the statement.

**Proposition 13.** *Let $c$ be a constant, let $p := c/n$ and let us consider the (1+1) $EA_c$. If $g$ is a linear function with monotone weights such that $E[\Delta_c(\ln(1+g), f, x)] \geq 0$ for all $f$ and $x$ as in Theorem 12, the following holds. For every $i \in [n]$, we have*

$$\ln(1 + g_i) \geq \max\left(\ln(2), \sum_{j=1}^{i-1} \binom{i-1}{j}(1-p)^{i-1-j}p^j \ln(1+j)\right)$$

*Proof.* For fixed $i$, let $f =$ BINVAL and $x = e_i$. As in the proof of Proposition 9 let $Y$ denote the mutation vector and let $A$ be the event that $f(x \oplus Y) > f(x)$. Then $A$ occurs if and only if $Y_i = 1$ and $Y_j = 0$ for all $j > i$.

$$0 \leq E[\Delta_c(\ln(1+g), f, x)] \leq E[\ln(1+g(x)) - \ln(1 + g(x \oplus Y)) \mid A]\Pr[A],$$

where $\Pr[A]$ has shown to be positive in the proof of Proposition 9. Therefore, $0 \leq E[\ln(1 + g(x)) - \ln(1 + g(x \oplus Y)) \mid A]$.

Given that $Y_i = 1$ and $Y_j = 0$ for all $j > i$ the first $i - 1$ bits are subject to independent, random mutation with mutation probability $p$. Thus, the probability that $k \leq i-1$ of the first $i-1$ bits flip equals $\binom{i-1}{k}(1-p)^{i-1-k}p^k$. Thus, considering the fact that $g_j \geq 1$ for all $j \in [n]$, we obtain

$$0 \leq \ln(1 + g_i) - \sum_{j=0}^{i-1} \binom{i-1}{j}(1-p)^{i-1-j}p^j \ln(1+j).$$

$\square$

**Proposition 14.** *Let $c$ be a constant and let us consider the (1+1) $EA_c$. Let $g$ be a linear function with monotone weights such that $E[\Delta_c(\ln(1+g), f, x)] \geq 0$ for all linear functions $f$ and every $x \in \{0,1\}^n$. If we set $p := c/n$, then*

$$\sum_{j=1}^{n} \ln(1 + g_j) \leq \ln(2)\frac{1 + p(n-1)}{p}.$$

*Proof.* Like in Proposition 11, let $f =$ ONEMAX and $x = e_1$. Then, the same arguments used there yield

$$0 \leq (1-p)\ln(1 + g_1) + p\sum_{j=2}^{n}\left(\ln(1+g_1) - \ln(1+g_j)\right).$$

The statement follows from resorting and $g_1 = 1$. $\square$

*Proof of Theorem 12.* Propositions 13 and 14 yield

$$\sum_{i=1}^{n}\sum_{j=1}^{i-1}\binom{i-1}{j}(1-p)^{i-1-j}p^j \ln(1+j) \leq \sum_{j=1}^{n}\ln(1+g_j) \leq \ln(2)\frac{1 + np - p}{p}.$$

We use Maple to compute that for $p = \frac{4}{100}$ the term on the left is bigger than 91, whereas the term on the right equals $124 \ln(2) \leq 86$. □

Note that we could improve the constant $c \geq 4$ in the previous proof if we took into account that for every $j \in [n]$ we have that $\ln(1 + g_i) \geq \ln(2)$. But, as written above, we do not elaborate this idea any further.

### 4.3 Distribution-Based Multiplicative Drift

A natural question to ask is whether the distribution-based approach of Jägerküpper [Jäg08] and in particular the application of Theorem 5 does help.

We show that this is not the case. More precisely, we show that there exist probability distributions satisfying the requirements of Theorem 5 which do not allow universal drift functions for $c \geq 7$.

To formulate this statement rigorously, we introduce the notion of $\Delta_c(f, g, \mathcal{D})$. In the style of definition 7, it denotes the change in the potential function $g$ of the (1+1) $\text{EA}_c$ minimizing the function $f$ with individuals distributed according to distribution $\mathcal{D}$.

**Definition 15** ($\Delta_c(f, g, \mathcal{D})$)**.** *Let $n \in \mathbb{N}$ and let $c$ be a constant. Moreover, let $f$ and $g$ be two functions on $\{0,1\}^n$ and $\mathcal{D} \colon \{0,1\}^n \to [0,1]$ be a probability distribution on $\{0,1\}^n$. Finally, let $x \in \{0,1\}^n$ be drawn according to $\mathcal{D}$ and let $y \in \{0,1\}^n$ be sampled by flipping each bit of $x$ independently with probability $c/n$. Then the random variable $\Delta_c(f, g, \mathcal{D})$ is defined as*

$$\Delta_c(f, g, \mathcal{D}) = \begin{cases} g(x) - g(y) & \text{if } f(y) \leq f(x), \\ 0 & \text{otherwise.} \end{cases}$$

We now show that in the setting of Theorem 5 linear universal drift functions do not exist for $c \geq 7$.

**Theorem 16.** *Let $n \in \mathbb{N}$ be sufficiently large, $c \geq 7$ a constant, and let $g \colon \{0,1\}^n \to \mathbb{R}$ be a linear function with monotone weights $1 = g_1 \leq \cdots \leq g_n$. Then there exist a linear function $f \colon \{0,1\}^n \to \mathbb{R}$ and a probability distribution $\mathcal{D} \colon \{0,1\}^n \to [0,1]$ with*

$$\Pr_{\mathcal{D}}[x_1 = 0] \leq \cdots \leq \Pr_{\mathcal{D}}[x_n = 0] \quad (5)$$

*such that* $\mathrm{E}[\Delta_c(f, g, \mathcal{D})] < 0$.

The rest of this section is devoted to the proof of the previous theorem. For this purpose, we consider the following collection of distributions on $\{0,1\}^n$.

**Definition 17** (Distributions $\mathcal{D}_k$ on $\{0,1\}^n$)**.** Let $n \in \mathbb{N}$ and $k \in [n]$. We define a distribution $\mathcal{D}_k \colon \{0,1\}^n \to [0,1]$ by setting for all $x \in \{0,1\}^n$

$$\mathcal{D}_k(x) := \begin{cases} 1/k & \text{if } x = e_i \text{ with } i \in [k], \\ 0 & \text{otherwise.} \end{cases}$$

Let $n \in \mathbb{N}$ be sufficiently large and assume that Theorem 16 does not hold. Then there exist a constant $c \geq 7$ and a linear function $g \colon \{0,1\}^n \to \mathbb{R}, x \mapsto \sum_{j=1}^n g_j x_j$ with weights $1 = g_1 \leq \cdots \leq g_n$ such that $\mathrm{E}[\Delta_c(f, g, \mathcal{D}_k)] \geq 0$ for all linear functions $f$ and for every $k \in [n]$.

The following Proposition gives an upper bound for the sum of the weights of $g$. It is a direct consequence of Proposition 11 for $\mathcal{D} = \mathcal{D}_1$.

**Proposition 18.** *Let $c$ be a constant. If $g$ is a linear universal drift function with monotone weights such that $\mathrm{E}[\Delta_c(f, g, \mathcal{D}_k)] \geq 0$ for all linear functions $f$ and for every $k \in [n]$, then $\sum_{i=1}^n g_i \leq n - 1 + n/c$.*

*Proof.* The proof is similar to the one of Proposition 11. Let $f = \textsc{OneMax}$. As $\mathcal{D}_1$ is the simple distribution with $\mathcal{D}_1[e_1] = 1$, we immediately have $x = e_1$ if $x$ is sampled from $\{0,1\}^n$ according to $\mathcal{D}_1$. Furthermore we have required that $1 \leq g_1 \leq \ldots \leq g_n$. That is, we are in exactly the same situation as in the proof of Proposition 11 and conclude the proof as elaborated there. □

A lower bound of the weights is given by the following result.

**Proposition 19.** *Let $c$ be a constant and let $g$ be a linear function with $1 = g_1 = \min_{j \in [n]} g_j$ and $\mathrm{E}[\Delta_c(f, g, \mathcal{D}_k)] \geq 0$ for all linear functions $f$ and for every $k \in [n]$. Furthermore, let $s = \min\{i \in \mathbb{N} \mid (1 - c/n)^i < 1/2\}$. For all $k \in [n]$ it holds that*

$$g_k \geq k - s - \tfrac{n}{c} + g_{k-1}(s + 1 - 2\tfrac{n}{c}).$$

*Proof.* Let $k \in [n]$. We set $f := \textsc{BinVal}$. Let $x$ be sampled from $\{0,1\}^n$ according to $\mathcal{D}_k$. Let $Y \in \{0,1\}^n$ with $Y_j = c/n$ independently for all $j \in [n]$. Again we abbreviate $p := c/n$. By the definition of $\mathcal{D}_k$ we have for all $i \leq k$ that $\Pr[x = e_i] = 1/k$. Thus,

$$0 \leq \mathrm{E}[\Delta_c(f, g, \mathcal{D}_k)]$$
$$= \sum_{i=1}^k \frac{1}{k} \Pr[f(e_i \oplus Y) \leq f(e_i)] \, \mathrm{E}[g(e_i) - g(e_i \oplus Y) \mid f(e_i \oplus Y) \leq f(e_i)].$$

As outlined in the proof of Proposition 9 it holds that $f(e_i \oplus Y) \leq f(e_i)$ if and only if either $Y = 0$ –in which case $g(e_i) - g(e_i \oplus Y) = 0$– or if both

$Y_i = 1$ and $Y_j = 0$ for all $j > i$ – in which case $f(e_i \oplus Y) < f(e_i)$. Thus,

$$0 \leq \sum_{i=1}^{k} \frac{1}{k} \Pr[f(e_i \oplus Y) < f(e_i)] \operatorname{E}[g(e_i) - g(e_i \oplus Y) \mid f(e_i \oplus Y) < f(e_i)]$$

$$= \frac{1}{k} \sum_{i=1}^{k} p(1-p)^{n-i} \operatorname{E}[g(e_i) - g(e_i \oplus Y) \mid f(e_i \oplus Y) < f(e_i)].$$

As in the proof of Proposition 9 we obtain that

$$\operatorname{E}[g(e_i) - g(e_i \oplus Y) \mid f(e_i \oplus Y) < f(e_i)] = \left(g_i - p \sum_{j=1}^{i-1} g_j\right).$$

Putting everything together we have

$$0 \leq \frac{1}{k} p \sum_{i=1}^{k} (1-p)^{n-i} \left(g_i - p \sum_{j=1}^{i-1} g_j\right)$$

$$= \frac{1}{k} p \left[(1-p)^{n-k} g_k + \sum_{i=1}^{k-1} g_i \left((1-p)^{n-i} - p \sum_{j=i+1}^{k} (1-p)^{n-j}\right)\right].$$

Multiplication by $kp^{-1}(1-p)^{k-n}$ and sorting yields

$$g_k \geq \sum_{i=1}^{k-1} g_i \left(p \sum_{j=i+1}^{k} (1-p)^{k-j} - (1-p)^{k-i}\right)$$

$$= \sum_{i=1}^{k-1} g_i \left(p \sum_{j=0}^{k-i-1} (1-p)^j - (1-p)^{k-i}\right)$$

$$= \sum_{i=1}^{k-1} g_i \left(p \frac{1 - (1-p)^{k-i}}{p} - (1-p)^{k-i}\right)$$

$$= \sum_{i=1}^{k-1} g_i (1 - 2(1-p)^{k-i}). \tag{6}$$

By definition of $s$ the summands in (6) are positive if and only if $k - i \geq s$. Thus, we can split the sum into a positive and a negative part. This yields

$$g_k \geq \sum_{i=1}^{k-s} g_i(1 - 2(1-p)^{k-i}) + \sum_{i=k-s+1}^{k-1} g_i(1 - 2(1-p)^{k-i}).$$

We now make use of the fact that $1 \leq g_i$ (on the left-hand side) and that the $g_i$ are monotonically increasing (on the right-hand side). This yields

$$g_k \geq \sum_{i=1}^{k-s}(1-2(1-p)^{k-i}) + \sum_{i=k-s+1}^{k-1} g_{k-1}(1-2(1-p)^{k-i})$$

$$= k - s - 2\sum_{i=s}^{k-1}(1-p)^i + g_{k-1}\big(s-1-2\sum_{i=1}^{s-1}(1-p)^{k-i}\big)$$

$$= k - s - 2\big(\tfrac{1-(1-p)^k}{p} - \tfrac{1-(1-p)^s}{p}\big) + g_{k-1}\big(s-1-2\big(\tfrac{1-(1-p)^s}{p}-1\big)\big)$$

$$= k - s - 2\tfrac{n}{c}((1-p)^s - (1-p)^k) + g_{k-1}(s+1-2\tfrac{n}{c}(1-(1-p)^s)).$$

Together with $(1-p)^s - (1-p)^k \leq \tfrac{1}{2}$ and $(1-(1-p)^s) \leq 1$, we obtain the desired $g_k \geq k - s - \tfrac{n}{c} + g_{k-1}(s+1-2\tfrac{n}{c})$. □

Note that $(1-\tfrac{c}{n})^{n/c} \leq e^{-1} < \tfrac{1}{2}$, which, by definition of $s$, results in $s \leq \tfrac{n}{c}$. Hence, $s + 1 - 2\tfrac{n}{c} < 0$. Therefore, the lower bound of $g_k$ provided by Proposition 19 is better, the smaller $g_{k-1}$. On the other hand, we know that the weights $g_i$ of $g$ are increasing in $i$. The idea of proving Theorem 16 is simply to use the better of the two estimates for one particular $g_k$.

*Proof of Theorem 16.* We use the upper and lower bound for $\sum_{i=1}^n g_i$ obtained in the previous two propositions and show that they contradict each other for $c \geq 7$. To this end, let us abbreviate $\ell := \lceil 6\tfrac{n}{c} \rceil$ and make a case distinction with respect to the size of $g_\ell$. First, let us assume that $g_\ell < 2$. Recall that $s + 1 - 2\tfrac{n}{c} < 0$. We can thus apply Proposition 19 to obtain

$$g_{\ell+1} \geq 6\tfrac{n}{c} + 1 - s - \tfrac{n}{c} + 2(s+1-2\tfrac{n}{c}) \geq \tfrac{n}{c} + s + 3.$$

Due the fact that $g_i \geq 1$ for all $i$, we can bound the sum of the weights of $g$ from below by

$$\sum_{i=1}^n g_i \geq (n-1) + \tfrac{n}{c} + s + 3\,.$$

But this inequality contradicts the upper bound $\sum_{i=1}^n g_i \leq (n-1) + \tfrac{n}{c}$ obtained in Proposition 18.

Therefore it must hold that $g_\ell \geq 2$. In this case, the monotonicity condition of the weights $g_1 \leq \ldots \leq g_n$ implies that $g_i \geq 2$ for all $i \geq \ell$. By definition we also have $g_j \geq 1$ for all $j \in [\ell-1]$. Thus,

$$\sum_{i=1}^n g_i \geq \lfloor 6\tfrac{n}{c} \rfloor + 2(n - \lfloor 6\tfrac{n}{c} \rfloor) \geq 2n - 6\tfrac{n}{c},$$

again contradicting $\sum_{i=1}^n g_i \leq (n-1) + \tfrac{n}{c}$ for $c \geq 7$. □

## 5 Conclusion

In this work we considered the state-of-the-art proof techniques for analyzing the runtime of the (1+1) EA optimizing linear functions. We found that both the classical proof via additive drift as well as the more recent multiplicative method stop working for mutation probabilities beyond $c/n$, where $c$ is a small constant. This problem cannot be solved by defining the weights $g_i$ of the drift function differently — we have shown that for any choice of $g$ there is a linear function $f$ such that the drift $\mathrm{E}[\Delta(f,g,x)]$ is negative for some search point $x$.

We also showed that also the Jägersküpper method fails for mutation probabilities larger than $7/n$. This raises the question how the current, generally very successful drift methods can be used with larger mutation probabilities.

As can be easily seen, we did not put too much effort in optimizing the constants $c$. Although we do not know the minimum value of this constant, we find that already the presented values are frighteningly close to the most commonly used mutation probability of $1/n$.

A more challenging problem arising from this work, naturally, is to find methods that work for mutation probabilities larger than these barriers. As our analysis shows, here either the drift function has to be chosen individually for each objective function, or different classes of drift functions than those regarded by us have to be used. Both might, though, again lead to tedious calculations.

**Note added in proof:** Indeed, at the recent PPSN conference Doerr and Goldberg [DG10] managed to prove the $\Theta(n \log n)$ bound for arbitrary $c/n$ mutation probabilities by defining a drift function for each linear objective function $f$ and each constant $c$. This construction, however, is quite technical.